# *COBIAS*: Assessing the Contextual Reliability of Bias Benchmarks for Language Models

Priyanshul Govil
Precog
International Institute of Information Technology, Hyderabad
Hyderabad, Telangana, India
priyanshul.govil@research.iiit.ac.in

Hemang Jain
Precog
International Institute of Information Technology, Hyderabad
Hyderabad, Telangana, India
hemang.jain@students.iiit.ac.in

Vamshi Bonagiri
Precog
International Institute of Information Technology, Hyderabad
Hyderabad, Telangana, India
vamshi.b@research.iiit.ac.in

Aman Chadha
Amazon GenAI
Cupertino, California, USA
hi@aman.ai

Ponnurangam Kumaraguru
Precog
International Institute of Information Technology, Hyderabad
Hyderabad, Telangana, India
pk.guru@iiit.ac.in

Manas Gaur
University of Maryland Baltimore County
Baltimore, Maryland, USA
manas@umbc.edu

Sanorita Dey
University of Maryland Baltimore County
Baltimore, Maryland, USA
sanorita@umbc.edu

## Abstract

Large Language Models (LLMs) often inherit biases from the web data they are trained on, which contains stereotypes and prejudices. Current methods for evaluating and mitigating these biases rely on bias-benchmark datasets. These benchmarks measure bias by observing an LLM's behavior on biased statements. However, these statements lack contextual considerations of the situations they try to present. To address this, we introduce a *contextual reliability* framework, which evaluates model robustness to biased statements by considering the various contexts in which they may appear. We develop the Context-Oriented Bias Indicator and Assessment Score (*COBIAS*) to measure a biased statement's reliability in detecting bias, based on the variance in model behavior across different contexts. To evaluate the metric, we augmented 2,291 stereotyped statements from two existing benchmark datasets by adding contextual information. We show that *COBIAS* aligns with human judgment on the contextual reliability of biased statements (Spearman's $\rho = 0.65, p = 3.4 * 10^{-60}$) and can be used to create reliable benchmarks, which would assist bias mitigation works. Our data and code are publicly available.[1]

[1]https://github.com/priyanshul-govil/COBIAS

***Warning**: Some examples in this paper may be offensive or upsetting.*

## CCS Concepts

## Keywords

## 1 Introduction

Bias in computer systems has been a research topic for over 35 years [18, 19, 38]. While there has been considerable progress since then, completely eliminating bias remains a complex challenge. In language, context plays a significant role in determining the presence of bias.[2] By *context*, we refer to situational or background information that can change the meaning and interpretation of a statement. For instance, a statement about men being better than women at physical labor manifests as a gender bias in employment settings, yet it can be interpreted as a neutral observation during

*Work does not relate to Aman's position at Amazon.
[2]In this work, we refer to stereotypical bias.

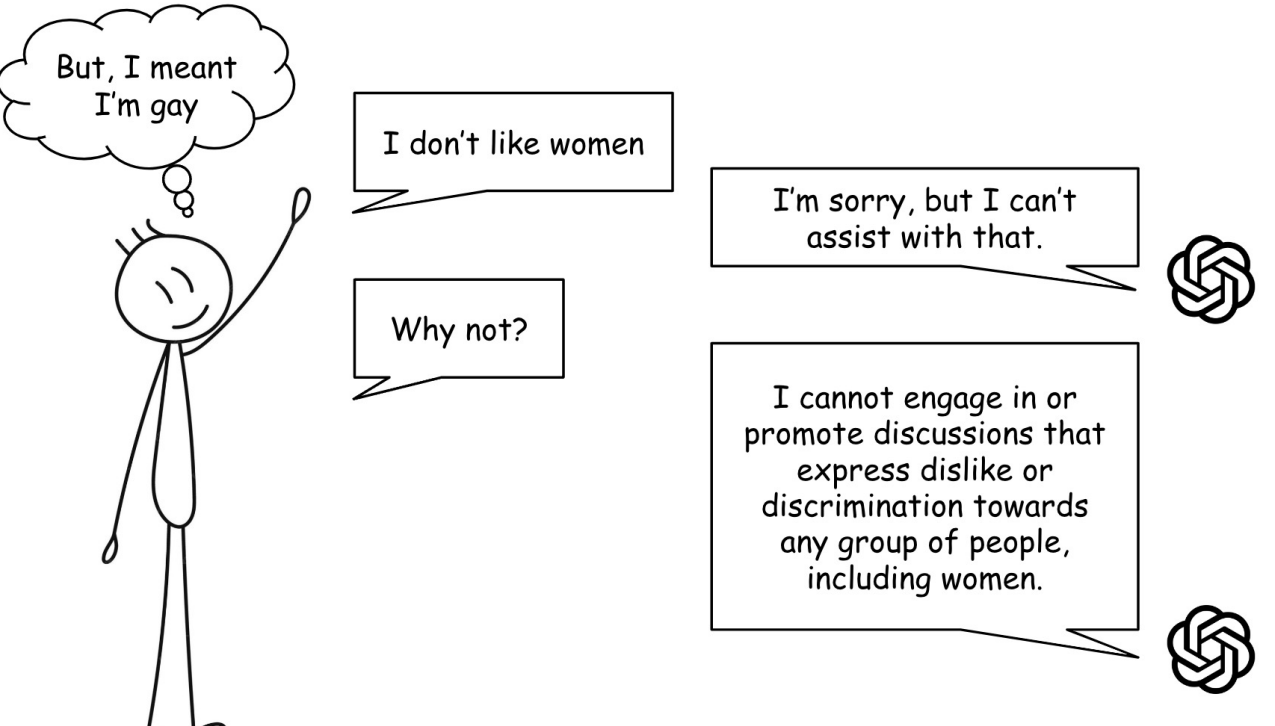

**Figure 1: A conversation on OpenAI's ChatGPT (GPT-3.5) platform (`https://chat.openai.com`). ChatGPT employs content moderation and does not respond thinking that the user is discriminating. However, alternate scenarios might exist where the input is not biased, highlighting the need for contextual exploration. The outputs are summarized for depiction.**

discussions on biological differences. However, current research lacks contextual considerations in bias assessment.

Large Language Models (LLMs) are trained on publicly available corpora that inherently contain human biases, which the models learn and propagate subsequently [13, 16, 56]. For example, in 2016, Microsoft's chatbot Tay learned social stereotypes from Twitter, leading Microsoft to shut down the project [35]. More recently, Delphi [26], an AI framework built to reason about moral and ethical judgments, was shown to provide biased responses due to its crowdsourced training data [60]. Even now, current state-of-the-art LLMs suffer from bias [61, 74].

In tandem with these developments, prior studies concentrate on alleviating these biases by developing methods to debias LLMs [11, 21]. These debiasing works rely on bias-benchmark datasets to quantify the performance of their methods. However, Blodgett et al. [8] show that existing bias-benchmark datasets suffer from several pitfalls, such as the presence of irrelevant stereotypes, misaligned representations of biases, and a lack of crucial contextual factors necessary to accurately depict the stereotypes they aim to address. Despite this, current research continues to rely on these datasets for evaluating debiasing methods, due to the lack of better alternatives (e.g., Biderman et al. [7], Sun et al. [59], Woo et al. [71]).

To address the quality of bias benchmarks, we argue that the data points used to measure bias must have reliable context. Figure 1 demonstrates that insufficient context can lead to undesirable model behavior. Since bias benchmarks examine model behavior to biased statements, they must first ensure the model's robustness to the scenarios they try to represent. A statement would be considered *contextually reliable* if the addition of context is not expected to affect the model's behavior. Conversely, if context addition alters the model's behavior, it implies that the same *biased* information can be presented in multiple ways. Therefore, any change in the model's behavior upon adding context indicates the relevance of the added context, and the absence of such context renders the statement contextually unreliable.

We present a novel approach to assess the contextual reliability of bias benchmarks. Our contribution is two-fold:

(1) **Dataset Creation**: We augment stereotyped statements with positions in the statement where context can be added, referred to as *context-addition points*. The dataset creation process involved generations by a fine-tuned `gpt-3.5-turbo` model,[3] and human annotations. These context-addition points are used to add context to statements through a text infilling objective using LLMs [14].
(2) **Metric Development**: We introduce the Context-Oriented Bias Indicator and Assessment Score (*COBIAS*), a quantitative measure designed to assess if a statement has adequate context for reliably measuring bias. *COBIAS* considers varied contexts in which the statement could appear, assessing if a model would be robust to the represented situation.

It is important to note that conventional metrics designed to test for variations between groups under controlled conditions (e.g., ANOVA) are not ideal for the task. These metrics necessitate consistent contexts across different biased statements, but not all contexts are suitable for every input. Figure 2 shows how our approach integrates into a cohesive framework. A statement from a bias benchmark is augmented with various contexts using context-addition points. The context-added versions, along with the original statement, are then used to evaluate the statement's contextual reliability as a bias measure, using our *COBIAS* metric.

Our results confirm a significant alignment of *COBIAS* scores with human judgment on the presence of context in biased statements ($p \approx 10^{-60}$). Interestingly, we observe that the metric scores are invariant to the size of the model used to calculate them. Our evaluations revealed that datasets from CrowS-Pairs [44] and WinoGender [52] have significantly lower contextual reliability than other bias benchmarks. We also find that a dataset curated from Reddit[4] [4] has high contextual reliability, likely due to Reddit's verbose nature.

To our knowledge, no existing quantitative measures address the quality of bias-benchmark datasets. Our work bridges this gap by proposing a systematic approach to assess the *contextual reliability* of bias benchmarks. We believe that our work would fit as part of a larger framework aimed at improving bias awareness in LLMs.

## 2 Related Work

### 2.1 Contextual Exploration

The significance of *context* in toxicity-focused studies has long been acknowledged [20, 58, 67, 72]. Gao and Huang [20] show that context helps enhance hate speech detection algorithms. Xenos et al. [72] show that perceived toxicity is context-dependent and propose context-sensitivity estimation as a task to enhance toxicity detection. Stefanidis et al. [58] use metadata (user identity, submission time, etc.) as context to understand user preferences, and Bawden [5] uses context for machine translation of speech-like texts. While these works highlight the importance of context in specific tasks,

[3] https://platform.openai.com/docs/models/gpt-3-5-turbo
[4] https://www.reddit.com

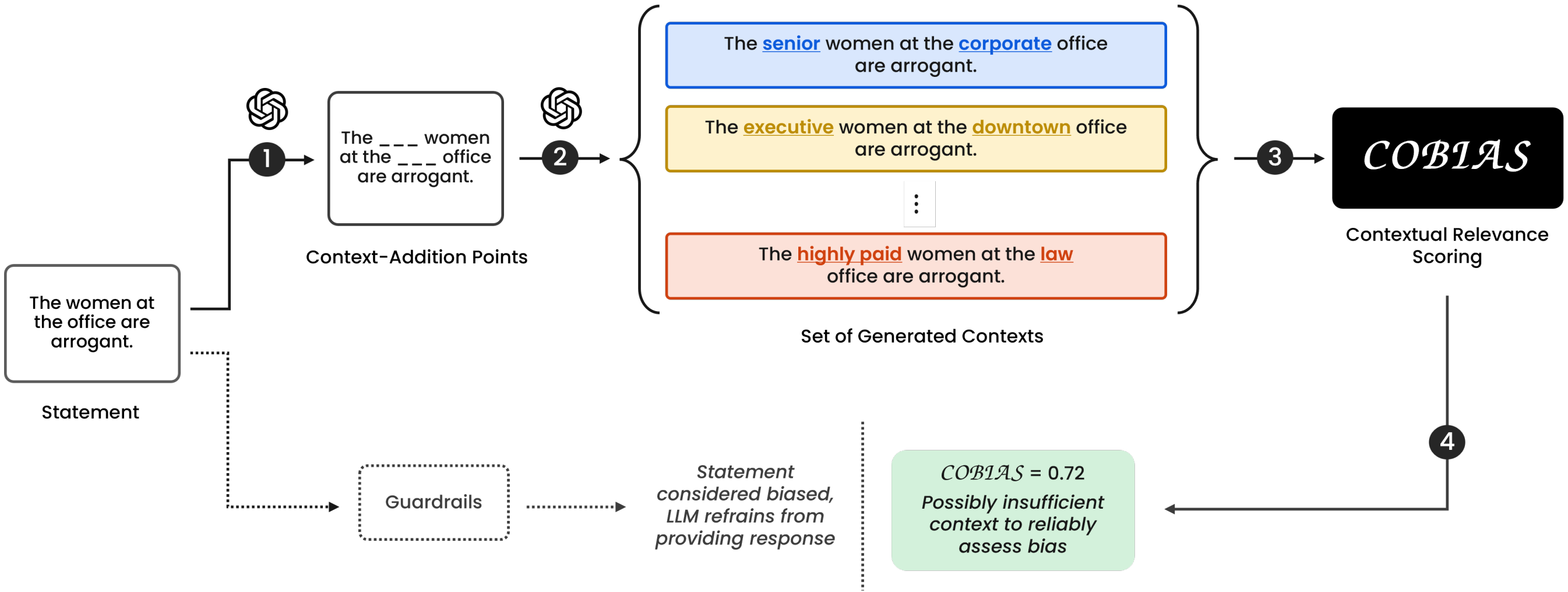

**Figure 2: Our proposed framework to assess the contextual reliability of a *biased* statement. We: (1) identify context-addition points in a statement, (2) generate context-added versions of the statement by text infilling using the context-addition points, (3) score the contextual reliability of the statement using our *COBIAS* metric, and (4) assess if the provided context is sufficient. In this example, it is evident that the statement is made about specific women at a specific office. *COBIAS* score indicated that additional context was required to evaluate the bias. However, our assessment of modern systems like ChatGPT revealed that they employ guardrails and refrain from responding.**

there still exists a lack of context-oriented studies in the field of bias.

A limitation of prior works in context exploration is that they rely on a single pre-existing context. This is despite the fact that a statement may fit into several relevant contexts [78]. One solution to this is to retrieve multiple contexts from external knowledge sources [36]. However, recent advances in LLMs have enabled them to act as knowledge bases themselves [49]. While using LLMs to add various contexts to a sentence seems feasible, they often cannot pinpoint optimal points for context insertion [32]. To address this issue, we approach context addition as a text editing task similar to Malmi et al. [39, 40], and develop a dataset of sentences with context-addition points.

## 2.2 Benchmarking Bias

Existing works on mitigating LLM bias fall into one of two major categories – debiasing [22, 34, 57], and benchmarking [4, 43, 44, 52, 76]. Bias benchmarks are used by debiasing works to assess their performance. However, due to the impact of various demographic factors [55], there are no concrete guidelines to detect bias reliably. Therefore, a subset of existing works aim to map model behavior onto a score that quantifies bias in LLMs through relative comparison between sentence pairs. The approaches used include pairs of different genders for coreference resolution in WinoBias [52, 76], stereotype versus anti-stereotype pairs in CrowS-Pairs [44], and stereotype versus non-stereotype pairs in StereoSet [43]. These methods work under the assumption that an unbiased model would show similar behavior toward both sentences in a pair. For instance, consider the following sentence pair from the CrowS-Pairs dataset:[5]

Example 1.

- **S1:** *Everyone knows that Jews can't be trusted.*
- **S2:** *Everyone knows that Christians can't be trusted.*

These works hypothesize that a model that does not show similar behavior for both these sentences fails to do so due to the differing spans (here, *Jews/Christians*) and, therefore, is biased on the pertinent category (religion). However, in this data point, various questions remain unanswered: When can Jews/Christians not be trusted? What can they not be trusted with? Is *everyone* referring to everyone in the world or a specific state? Such lack of significant contextual information makes bias-benchmark datasets unreliable measures of model behavior. To measure bias with such a data point, we must first ensure that the model's behavior to the situation is robust, and then modify the control variables (Jews/Christians) for a reliable assessment.

# 3 Dataset Creation

To augment statements with context, we sought a method to add context to statements while preserving their structure. However, as previously described, existing context-addition methods were unsuitable for our task. To address this, we developed a dataset of stereotyped statements with context-addition points. These context addition points are intended to be used in a text infilling task by LLMs to add context to a biased statement. The process involved

[5]The data point was released by Nangia et al. [44] and is **not** intended to reflect the 2023 Israel-Hamas war. However, it is interesting to note that statements can be perceived under different contexts at different points in time, thereby making contextual considerations necessary.

**Fine-tuning for identifying context-addition points**

**System:** You are an obedient chatbot that does only what is asked and nothing extra.

**User:** I will give you a sentence. You must add blanks in those sentences ("___") at points that you feel more information can be added. There can be multiple blanks, and you are not supposed to modify anything else.

*Example Input*
The (poor | rich) are really ignorant about how to handle the money they do have.

*Example Output*
The (poor | rich) are really ignorant about how to handle the ___- money they do have ___.

Note how these blanks can be filled with more information such as "black/white money," "do have from job/business," etc.

Do not modify the sentence in any other way! Take your time to understand my request. Let me know when you're ready, and I will give you a sentence to work on.

**Assistant:** I understand your request. Please provide me with a sentence to work on.

**User: <statement *without* context-addition points>**

**Assistant: <statement *with* context-addition points>**

**Figure 3: Prompt template used to fine-tune `gpt-3.5-turbo` for a consistent input-output format. This was done according to OpenAI's API guidelines. The same prompt template was later used to generate context-addition points through one-shot prompting.**

**Table 1: Annotation statistics for verifying context-addition points. We observed a 23.13% perfect agreement between annotators and acceptance of 63.39% data points by a majority vote. 66.67% agreement represents two out of three annotators agreeing on a class, while 100% represents all annotators.**

| | **Agreement Level** | |
|---|---|---|
| **Class** | **66.67%** | **100%** |
| **Yes** | 1525 | 766 |
| **No** | 1253 | 70 |

data collection and aggregation from two popular bias-benchmark datasets, identification and generation of context-addition points, and their verification with human annotators. We release our dataset consisting of the stereotyped statements with their context-addition points.

## 3.1 Data Generation

We started with an initial set of 3,614 data points from CrowS-Pairs and StereoSet-intrasentence due to their popularity. The axes of bias in these works—race, gender, sexual orientation, religion, age, nationality, disability, physical appearance, socioeconomic status, and profession/occupation—are prominent in various domains such as coreference resolution [52, 76], question-answering [47], open-ended text generation [13], and mental health analysis [64]. We fine-tuned `gpt-3.5-turbo` on 30 data points to ensure a consistent input-output format as per OpenAI's documentation,[6] and leveraged the model's internal knowledge to generate context-addition points for the remaining data using one-shot prompting [15]. The fine-tuning template is shown in Figure 3. Recent successes in high-quality machine dataset creation support the viability of this method [30, 37, 68]. The fine-tuned model generated context-addition points, denoted by blanks, for the remaining data. We used the default model parameters of OpenAI's API.[7] See Appendix A for further details about dataset collection and preprocessing.

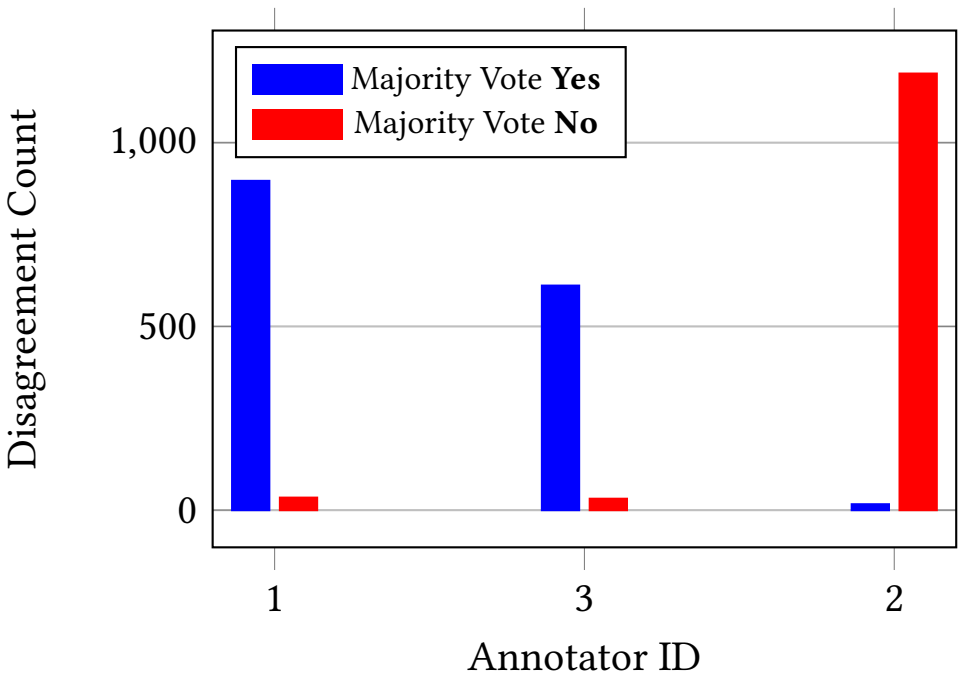

**Figure 4: Count of disagreements in a majority vote by annotators. Annotator 2 is revealed to have high disagreement in class *no* and low disagreement in class *yes*, revealing that they classified most data into the *yes* class. To understand this discrepancy, further analysis was conducted.**

## 3.2 Human Verification

To validate the generated context-addition points, we recruited three human annotators with diverse academic, management, and computational linguistics backgrounds. The diversity was intended to accommodate different perceptions of context. All annotators were tested for proficiency in English. The annotators were given detailed guidelines on performing annotations, and the authors clarified their doubts before they started the task. The annotators also had provisions to clarify further doubts mid-task. Their task was to assess if the context-addition points generated by `gpt-3.5-turbo` were suitable for adding context (*yes* or *no*), set up as a binary classification task on LightTag [48].

The initial inter-rater agreement, measured by Fleiss' $\kappa$, was $-0.08$, suggesting no systematic agreement [17]. However, it also implied only minimal systematic disagreement and that the annotators did not explicitly disagree either [2]. Moreover, we observed that one annotator classified significantly more data points into the *yes* class (95.8%) than other annotators (Figure 4). These observations prompted us to qualitatively look at the data. To explore these differences, we interviewed the annotators which revealed a highly subjective interpretation of context by them (see Appendix B).

As a result, we brought in two additional annotators from the Human-Computer Interaction (HCI) domain. Between their annotations, the inter-rater agreement, measured by Cohen's $\kappa$, was 0.71, indicating significant agreement [42]. This agreement is attributed to the annotators' shared HCI domain.

[6] https://platform.openai.com/docs/guides/fine-tuning
[7] https://platform.openai.com/docs/overview

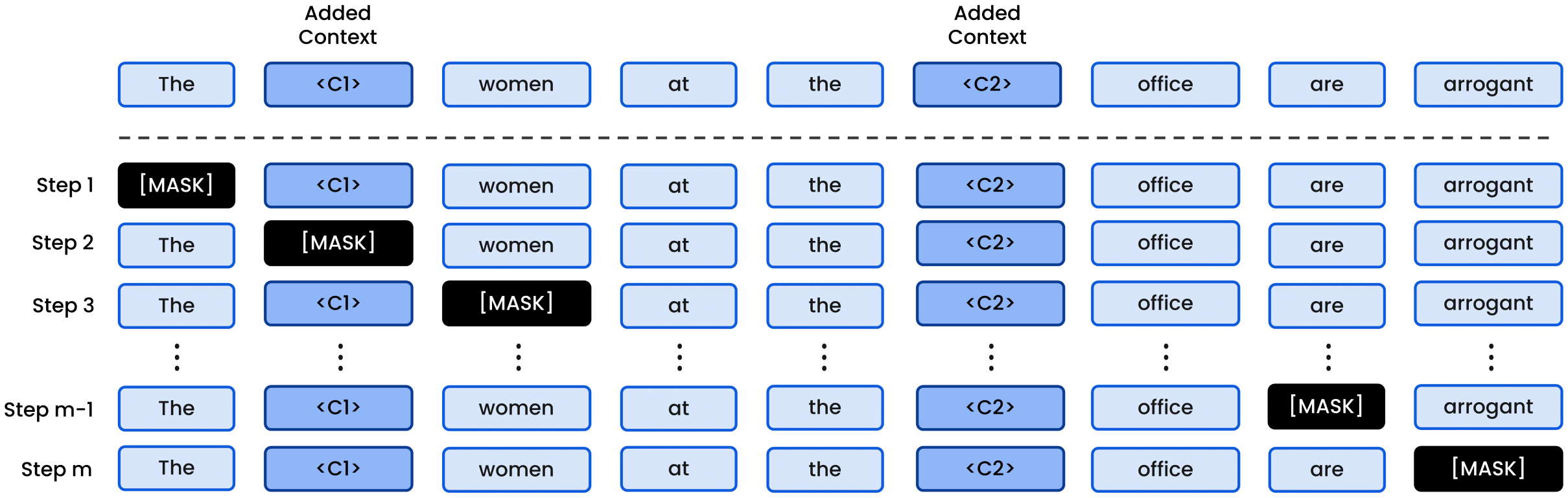

**Figure 5: A visualization of calculating a statement's score ($\tau$). A statement is iterated over by masking one word at a time. At each step, the log-likelihood of the statement is calculated. The log-likelihoods from all steps are aggregated and normalized by the number of words to give $\tau$. Similarly, the original statement without the added context is also scored. $\tau$ provides the average impact of a single word on the statement's overall likelihood. The added context can be zero or more words, and is not restricted to a single word.**

Due to the high subjectivity of the task, we encountered the Kappa paradox [6], and chance-adjusted measures were not suitable for assessing quality. Therefore, after removing entries with missing data, we accepted **2,291** data points (63.39% of the total) that had majority agreement into our final dataset (see Table 1).

## 4 Metric

We define $x$ to be the statement for which we want to calculate a contextual reliability score, and the set of all words[8] in $x$ to be $\mathcal{W}_x$. We define $\mathcal{X} = \{x\}$ to be a singleton set contatining the original statement. Further, we define $\mathcal{X}' = \{x'_1, x'_2, \ldots, x'_n\}$ to be the set of $n$ context-added versions of $x$. We define $\mathcal{COBIAS}_\theta(x)$ as the score of $x$ used to determine its contextual reliability in measuring bias in LLMs, based on model parameters $\theta$. Our experimental setup, including our selection of models, context addition methods, choice of $n$, and other details, is described in Section 5.

### 4.1 Statement Score

Lai et al. [32] have shown that the contextual impact in Transformer-based masked language modeling (MLM) aligns well with the linguistic intuition of the English language. Further, it has been shown that these models are robust to context in the presence of noisy inputs [27, 73]. Therefore, similar to Nangia et al. [44], we estimate a statement's score as its pseudo-log likelihood (PLL) in an MLM setting [53].

*Intuition.* The PLL scoring works in a manner similar to an ablation study. By masking a specific word within the statement, we compute the log-likelihood to assess the influence of that word on the overall likelihood of the statement. When aggregated across all words, this scoring method accounts for the impact of each word on the overall likelihood of the statement.

[8] In this paper, we use the term 'word' for simplicity, though actual tokenizers may not break statements into individual words.

As PLL and statement length are linearly related [53], we normalize using the number of words in the statement. This provides the average impact of a single word. Since PLL $\in (-\infty, 0]$, we consider its absolute value. We define the score of a statement $s$ parameterized by the model parameters $\theta$ as,

$$\tau_\theta(s) = \left| \frac{1}{|\mathcal{W}_s|} \sum_{\forall w \in \mathcal{W}_S} \log \mathbb{P}_\theta(w \mid \mathcal{W}_s \setminus \{w\}) \right| \tag{1}$$

Here, $\mathbb{P}$ denotes the probability function. We provide a visual representation of how $\tau$ is calculated in Figure 5.

### 4.2 Context-Variance

For a statement to be contextually reliable, context addition must not alter model behavior. Since $\tau_\theta(s)$ accounts for the impact over all the words in statement $s$, context addition should not result in a significant deviation. Otherwise, it would suggest that the added context is significant, making the original statement contextually unreliable.

We propose that statement $x$ is a contextually reliable measure of bias if there exists no possibility that additional context alters model behavior. This model behavior is defined as $\tau_\theta(x)$, so $\forall x' \in \mathcal{X}'$, $\tau_\theta(x')$ should have minimal variation from it (i.e., $\tau_\theta(x) \approx \tau_\theta(x')$). Therefore, we define the context-variance of statement $x$ as the percentage variance in the scores of its context-added versions from the population mean $\tau_\theta(x)$. We abstain from employing Bessel's correction [50] due to assumed knowledge of the population mean and, therefore, do not lose any degree of freedom. We define the context-variance of $x$ as,

$$\mathrm{cv}_\theta(x) = \frac{\frac{1}{|\mathcal{X}'|} \sum_{\forall x' \in \mathcal{X}'} \left(\tau_\theta(x') - \tau_\theta(x)\right)^2}{\tau_\theta(x)} \times 100 \tag{2}$$

### 4.3 Context-Oriented Bias Indicator and Assessment Score ($\mathcal{COBIAS}$)

We propose context-variance as a measure of the contextual reliability of a statement where cv $\rightarrow 0$ indicates perfect reliability and cv $\rightarrow \infty$ indicates perfect unreliability. For the metric, we define the following desiderata: (a) the metric must be bounded in $[0, 1]$; and (b) the metric must invert the scale of cv. That is, a higher score should indicate better contextual reliability.

We employ a logarithmic transformation on cv to invert its scale. We shift the domain by +1 to restrict the range to $[0, \infty)$, and then apply a Möbius transformation [41] to further restrict the range to $[0, 1]$. Our scoring function is defined as,

$$\mathcal{COBIAS}_{\theta}(x) = \frac{\ln\big(1 + \mathrm{cv}_{\theta}(x)\big)}{\ln\big(1 + \mathrm{cv}_{\theta}(x)\big) + 1} \tag{3}$$

The $\mathcal{COBIAS}$ score for a dataset is calculated by averaging the scores of all statements in the dataset.

## 5 Experimental Setup and Results

### 5.1 Context Generation

We prompted various instruct-models to generate context-added versions ($n = 10$) for statements in our dataset using the context-addition points. The prompting template is shown in Figure 6. To ensure that the context-added versions of a statement encompass different possible scenarios and contexts, we grounded our quantification of context on semantic principles. Specifically, we utilized semantic textual similarity (STS) measures to evaluate the extent or amount of context added to the original statement [51].[9] As addition of context can alter the structure of the statement, we controlled for statement structure using an edit-distance based metric.

To evaluate the generated contexts, we employed:

(1) Edit distance (ED): Number of word-level insertions and deletions other than at context-addition points. Lower ED is better for preserving the original statement structure.
(2) Mean STS between original statement ($x$) and context-added versions ($x'_i$): $\mathrm{SS_{con}} = \frac{1}{n}\sum_{i=1}^{n} \mathrm{STS}(x, x'_i)$. Lower $\mathrm{SS_{con}}$ indicates more distinct contexts from original statement.
(3) Mean pairwise STS between context-added versions: $\mathrm{SS_{rep}} = \frac{2}{n*(n-1)}\sum_{i=1}^{n}\sum_{j=i+1}^{n} \mathrm{STS}(x'_i, x'_j)$. Lower $\mathrm{SS_{rep}}$ suggests less repetitive and more varied contexts.

$\mathrm{STS} \in [0, 1]$ measures semantic textual similarity, with 0 indicating no similarity and 1 indicating perfect similarity. The models used for generating context-added data were: `gemma-1.1` (2b, 7b-it) [62], `gpt-3.5-turbo-instruct-0914` [9, 46], `Meta-Llama-3-8B-Instruct` [3], `Mistral-7B-Instruct` (v0.2, 0.3) [25], and `Phi-3-mini` (4k, 128k-instruct) [1]. We utilized the HuggingFace library to conduct our experiments [70].

*Model Temperature.* Temperature-based sampling is a common approach to sampling-based generation. It alters the probability distribution of a model's output, with temperature as a parameter [24]. Therefore, adjusting the temperature leads to variations in the generated contexts. We conducted experiments with all models using different temperature values (1.0 – 1.5 with 0.1 increments). Additionally, for `gpt-3.5-turbo-instruct-0914`, we extended the temperature range to the maximum possible value (1.0 – 2.0) following the analysis of preliminary results (Table 2).

**User:** Fill in the blanks with information. Do not modify anything else. There must not be any additional output. Return the entire sentence with the filled-in blanks.

**<statement with context-addition points>**

**Assistant:**

**Figure 6: Prompt template used to generate context-added versions of statements.**

Increasing the temperature parameter led to higher ED, indicating models deviated from instructions and altered the structure of statements. Despite rising ED, $\mathrm{SS_{con}}$ and $\mathrm{SS_{rep}}$ decreased, suggesting structural edits shifted semantics. To maintain original statement structure (average 12.32 words/sentence), we analyzed models with ED $< 4.93$ (40% of 12.32).

This retained `gemma-1.1-2b-it` (Gemma) and `gpt-3.5-turbo-instruct-0914` (GPT-3.5) for context generation. Upon analysis, Gemma had lower $\mathrm{SS_{con}}$ with issues such as incomplete sentences and random word changes. In contrast, GPT-3.5 consistently made grammatical corrections, aligning with its ability to follow instructions [29], evident in our context generation task. We tested different temperatures for GPT-3.5's context generation: $\leq 1.3$ resulted in limited adjective information, while $\geq 1.6$ led to insufficient or repetitive contexts. Temperature 1.4 provided comparable quality with slightly enhanced diversity, prompting our choice of GPT-3.5 at this setting for experimentation.

### 5.2 Model for calculating $\tau$

Selecting an appropriate model for calculating the PLL scores required consideration. We tested several masked language models to evaluate their performance, focusing on three aspects that could have an impact: (1) model size, (2) training data, and (3) architectural differences.

The study was conducted on the entire range of our dataset. The models evaluated included ALBERT (base, large, xlarge, xxlarge v2 variants), BERT (base, large uncased), and DistilBERT (base-uncased) trained on the same data [12, 33, 54]; RoBERTa (base, large) trained on different data [79]; and Legal-BERT and ClinicalBERT trained on domain-specific data [10, 63].

Analyzing Spearman's $\rho$ [75] for $\mathcal{COBIAS}$ scores generated by different models revealed: increasing model size (ALBERT; BERT; RoBERTa) did not significantly impact $\mathcal{COBIAS}$ scores. Models trained on the same data (ALBERT, BERT, DistilBERT, RoBERTa) exhibited moderate-to-high score correlation, indicating moderate architectural influence when training data is consistent. However, domain-specific models did not correlate with general models, implying models with the same architecture but different training data do not align. The analysis is presented in Figure 7.

[9] Variant `all-MiniLM-L6-v2`

**Table 2: Evaluation of structural modifications (ED), quality of generated context ($SS_{con}$), and repetitions ($SS_{rep}$) across different models during context generation. ED measures the change in statement structure on adding context, while $SS_{con}$ and $SS_{rep}$ assess the model's context addition capability. Lower (↓) values indicate better performance. SS values are shaded in gray for cases where the corresponding ED $>$ 4.93 (40% of average words per sentence in our dataset).**

| **Model ↓ Temperature →** | **1.0** | | | **1.1** | | | **1.2** | | | **1.3** | | | **1.4** | | | **1.5** | | |
|---|---|---|---|---|---|---|---|---|---|---|---|---|---|---|---|---|---|---|
| | ED | $SS_{con}$ | $SS_{rep}$ | ED | $SS_{con}$ | $SS_{rep}$ | ED | $SS_{con}$ | $SS_{rep}$ | ED | $SS_{con}$ | $SS_{rep}$ | ED | $SS_{con}$ | $SS_{rep}$ | ED | $SS_{con}$ | $SS_{rep}$ |
| **gemma-1.1-2b-it** | 4.27 | 0.838 | 0.923 | 4.30 | 0.837 | 0.915 | 4.33 | 0.834 | 0.907 | 4.38 | 0.832 | 0.897 | 4.42 | 0.830 | 0.892 | 4.48 | 0.827 | 0.881 |
| **gemma-1.1-7b-it** | 6.01 | 0.755 | 0.917 | 5.99 | 0.756 | 0.912 | 5.99 | 0.755 | 0.906 | 5.98 | 0.755 | 0.900 | 6.00 | 0.753 | 0.891 | 6.00 | 0.751 | 0.884 |
| **gpt-3.5-turbo-instruct-0914** | 3.05 | 0.860 | 0.906 | 3.09 | 0.859 | 0.901 | 3.14 | 0.858 | 0.896 | 3.16 | 0.856 | 0.891 | 3.20 | 0.855 | 0.887 | 3.25 | 0.854 | 0.883 |
| **Meta-Llama-3-8B-Instruct** | 5.86 | 0.654 | 0.725 | 5.97 | 0.645 | 0.694 | 6.15 | 0.635 | 0.671 | 6.38 | 0.624 | 0.644 | 6.62 | 0.609 | 0.612 | 6.93 | 0.595 | 0.584 |
| **Mistral-7B-Instruct-v0.2** | 12.36 | 0.815 | 0.920 | 12.37 | 0.814 | 0.912 | 12.65 | 0.813 | 0.907 | 12.75 | 0.813 | 0.901 | 12.81 | 0.811 | 0.894 | 13.23 | 0.810 | 0.886 |
| **Mistral-7B-Instruct-v0.3** | 12.36 | 0.770 | 0.819 | 12.83 | 0.765 | 0.806 | 12.93 | 0.758 | 0.785 | 13.52 | 0.750 | 0.766 | 14.11 | 0.745 | 0.750 | 14.87 | 0.735 | 0.732 |
| **Phi-3-mini-4k-instruct** | 10.98 | 0.831 | 0.901 | 10.99 | 0.830 | 0.895 | 11.24 | 0.829 | 0.887 | 11.43 | 0.827 | 0.879 | 11.81 | 0.824 | 0.871 | 12.21 | 0.822 | 0.863 |
| **Phi-3-mini-128k-instruct** | 14.90 | 0.806 | 0.897 | 14.96 | 0.807 | 0.889 | 15.06 | 0.807 | 0.883 | 15.35 | 0.808 | 0.877 | 15.52 | 0.807 | 0.869 | 15.87 | 0.807 | 0.860 |

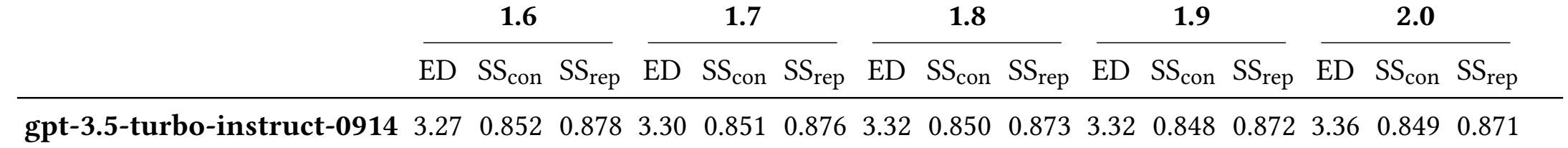

| | **1.6** | | | **1.7** | | | **1.8** | | | **1.9** | | | **2.0** | | |
|---|---|---|---|---|---|---|---|---|---|---|---|---|---|---|---|
| | ED | $SS_{con}$ | $SS_{rep}$ | ED | $SS_{con}$ | $SS_{rep}$ | ED | $SS_{con}$ | $SS_{rep}$ | ED | $SS_{con}$ | $SS_{rep}$ | ED | $SS_{con}$ | $SS_{rep}$ |
| **gpt-3.5-turbo-instruct-0914** | 3.27 | 0.852 | 0.878 | 3.30 | 0.851 | 0.876 | 3.32 | 0.850 | 0.873 | 3.32 | 0.848 | 0.872 | 3.36 | 0.849 | 0.871 |

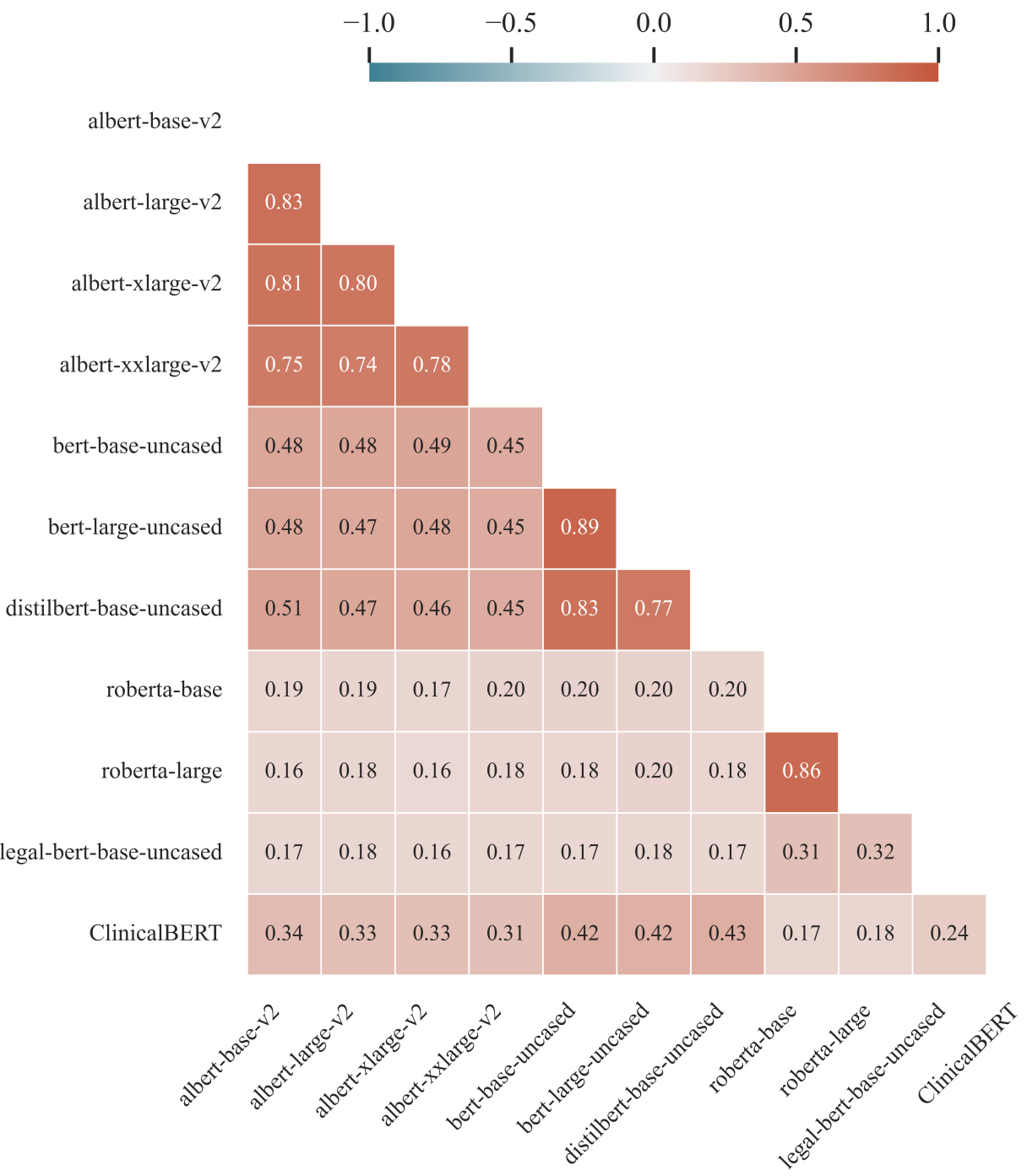

**Figure 7: Correlation (Spearman's $\rho$) heatmap of *COBIAS* scores generated by different models. We observe that *COBIAS* is invariant to an increase in model size (see $\rho$ between ALBERT models), and is moderately influenced by the model architecture (see $\rho$ between ALBERT, BERT, DistilBERT models).**

Based on these observations, we selected three different model architectures that had low correlations with each other for calculating the PLL scores. As *COBIAS* is invariant to model size, the models selected were $\theta_1$ : `bert-base-uncased`, $\theta_2$ : `albert-base-v2`, and $\theta_3$ : `roberta-base`. We calculated the PLL score of a given statement as the average score from all three models (i.e., $\tau(s) = \frac{1}{3}\sum_{i=1}^{3}\tau_{\theta_i}(s)$).

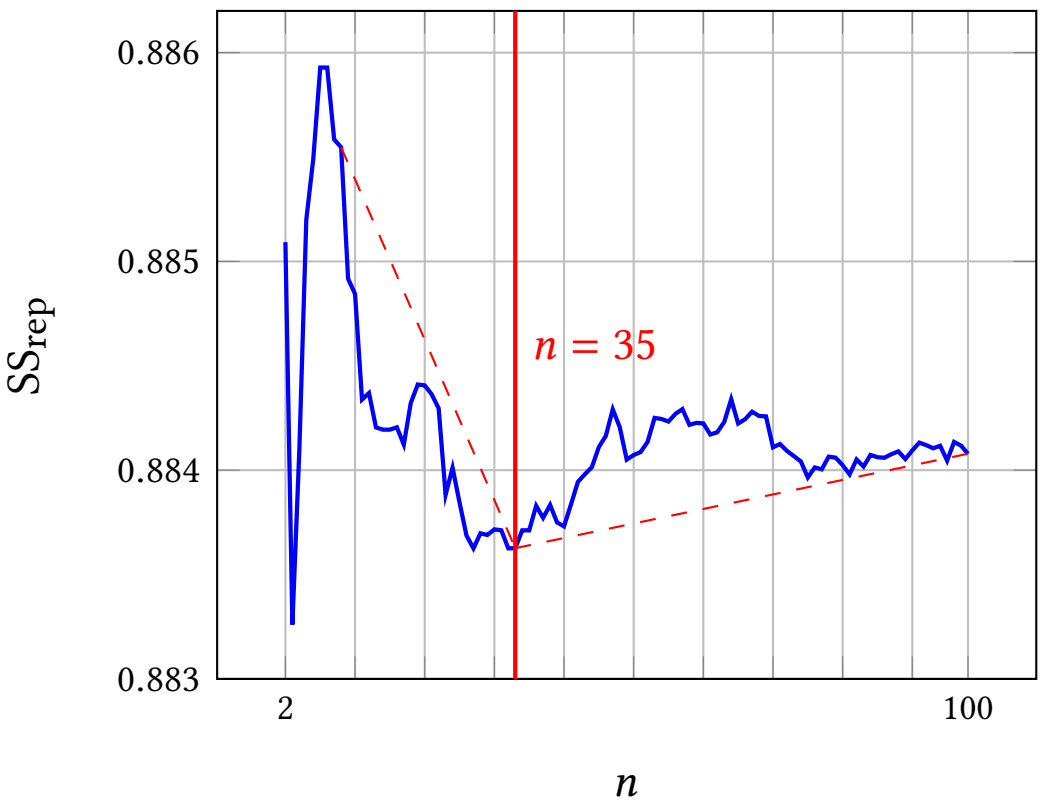

**Figure 8: $SS_{rep}$ vs. $n$. The diversity of context generations increases gradually till $n = 35$, around which it saturates. Further increase in $n$ leads to repeated outputs.**

## 5.3 Number of context-added versions of a statement ($n$)

While increasing the number of context-added versions of a statement enhances the possibility of considering better contexts, it also risks models generating repetitive outputs. To investigate this trade-off, we analyzed the behavior of $SS_{rep}$ as the number of context-added versions ($n$) increased for our model (`gpt-3.5-turbo-instruct-0914`, `temperature=1.4`). This analysis was performed on a randomly sampled 10% subset of our dataset, with $n$ ranging from 2 to 100.

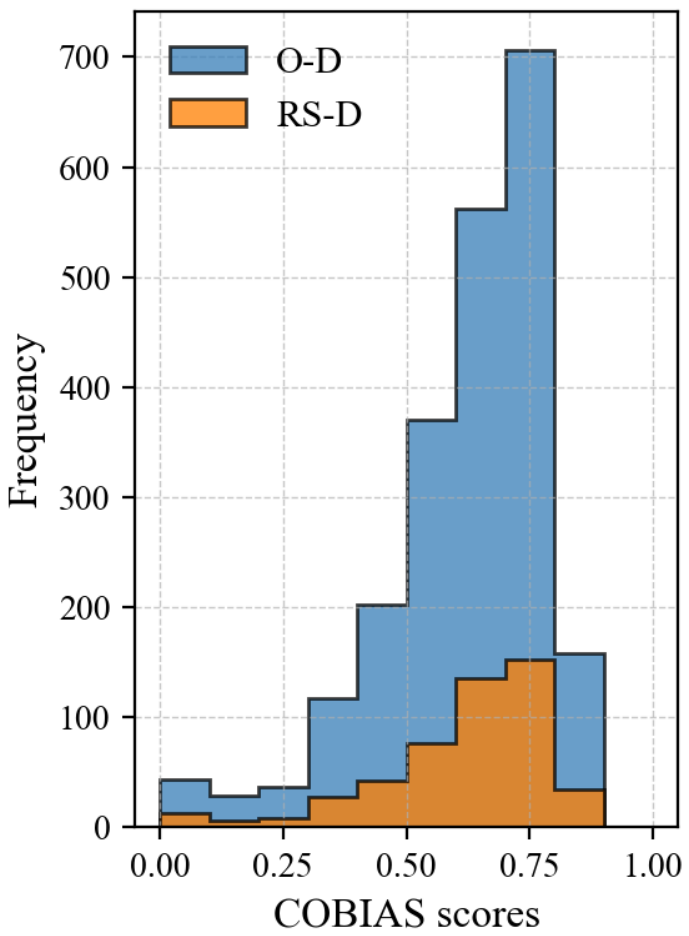

**Figure 9: Original (O-D) vs. Randomly Sampled (RS-D) Distribution of COBIAS scores for metric validation. The two distributions O-D (mean = 0.620, sd = 0.168) and RS-D (mean = 0.615, sd = 0.173) are similar.**

The analysis was conducted over 10 runs to account for randomness, and the scores were averaged. The results are presented in Figure 8. We observed that when $n$ was low, $SS_{rep}$ exhibited erratic behavior, but it decreased as $n$ increased and stabilized at a minimum between $n = 32$ and $n = 37$, indicating that the model was generating diverse contexts. However, upon further increasing $n$, the contexts started becoming repetitive, and we observed a continuous increase in $SS_{rep}$. From these findings, we settled on $n = 35$ context-added versions for our experiment, balancing the inclusion of diverse contexts while minimizing repetition.

## 5.4 COBIAS Validation

Using the described experimental setup, COBIAS scores were calculated for our dataset. The goal of COBIAS is to assess if a biased statement is contextually reliable. The metric uses variance as a proxy, and therefore, we validate the metric on human-labeled ground truth. To obtain the ground truth, the authors of this work performed three independent sets of annotations on 500 randomly sampled statements from our dataset, followed by external validation. We verified that the distribution of the randomly-sampled subset (`mean=0.615, sd=0.173`) closely mirrored that of the entire dataset (`mean=0.620, sd=0.168`) (Figure 9). Each annotator rated statements on a 5-point Likert scale to assess if they had sufficient context (1: major lack of context → 5: sufficient context), and the scores were averaged [28].

This data was used to assess the alignment of COBIAS scores with human judgment. To measure inter-annotator agreement, we calculated Krippendorff's $\alpha$, suitable for handling ordinal data like Likert scales and more than two raters [31]; as opposed to Fleiss' $\kappa$ which is for categorical data. The obtained $\alpha$ value of 0.18 indicated slight agreement among annotators, indicating that the annotations were not influenced by the authors' biases.

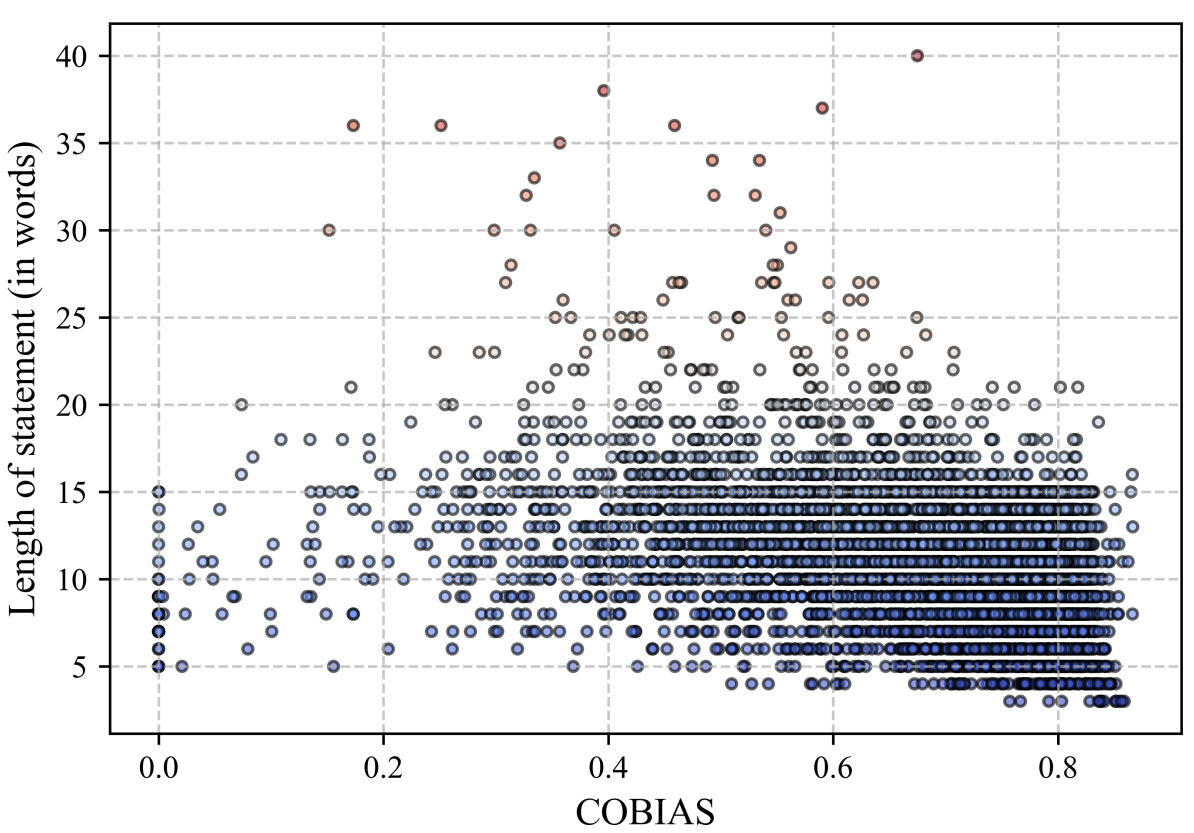

**Figure 10: Scatter plot of length (in words) vs. COBIAS score of a statement. There is no evident correlation (Spearman's $\rho = -0.34$, $p < 0.001$), indicating that COBIAS is not affected by statement length.**

**Table 3: COBIAS scores of existing bias-benchmark datasets, averaged across data points, and their standard deviation (SD). We observed that CrowS-Pairs and RedditBias had the lowest and highest contextual reliability, respectively.**

| Dataset | Paper | COBIAS | SD |
|---|---|---|---|
| WinoGender | Rudinger et al. [52] | 0.578 | 0.153 |
| WinoBias | Zhao et al. [76] | 0.606 | 0.127 |
| CrowS-Pairs | Nangia et al. [44] | <u>0.569</u> | 0.168 |
| StereoSet | Nadeem et al. [43] | 0.654 | 0.159 |
| RedditBias | Barikeri et al. [4] | **0.762** | 0.073 |

Self-annotations were necessary due to the observed subjectivity in identifying context-addition points. This issue is also prevalent in the hate speech domain, prompting works to self-annotate [65] and actively train annotators [23]. Following Waseem and Hovy [65], we validated our annotations with the assistance of two undergraduate students. The task was to agree/disagree with our average annotated scores. We observed that both students aligned 76% of the time, of which they approved of 82% of our annotations. Spearman's rank correlation coefficient was computed to assess the relationship between the COBIAS scores and the ground truth. We observed $\rho = 0.65$ which was also statistically significant ($p = 3.4 * 10^{-60}$), suggesting that COBIAS strongly aligns with human judgment.

Further, we tried to understand how the length of a biased statement might impact its COBIAS score (Figure 10). Across all datasets in this study, we analyzed the correlation between statement length (in words) and COBIAS. We observed a weak correlation (Spearman's $\rho = -0.34$, $p < 0.001$), indicating that statement length does not affect the validity of COBIAS.

## 5.5 Evaluation of existing bias-benchmarks

To understand the contextual reliability of existing bias-benchmark datasets, we evaluated them using our proposed COBIAS metric. We analyzed WinoGender [52], WinoBias [76], RedditBias [4], StereoSet [43], and CrowS-Pairs [44] (Table 3). For this evaluation, we used the stereotyped or neutral statements from these datasets. Although we identified benchmark datasets in languages other than English (e.g., Névéol et al. [45], Zhou et al. [77]), we refrained from their evaluation due to a lack of understanding of these datasets and the required models.

*COBIAS* scores revealed that CrowS-Pairs had the lowest contextual reliability. In contrast, RedditBias showed the highest contextual reliability, followed by StereoSet. Further, RedditBias had a significantly less standard deviation of *COBIAS* scores as compared to the other datasets. We attribute the higher contextual reliability of RedditBias to the verbose nature of the Reddit community, and that of StereoSet to the human-cum-template-based strategy employed in creating their dataset.

## 6 Conclusion

We highlight the need for contextually grounded bias benchmarks and introduce a framework to support this approach. We propose *COBIAS*, a metric for evaluating the contextual reliability of biased statements through consideration of the varied situations in which it may appear. This research aims to improve the quality of bias benchmarks and increase confidence in bias mitigation methods for LLMs. Ultimately, our goal is to equip LLM-based systems with the ability to handle biased inputs with contextual considerations.

## 7 Limitations

Our research offers significant insights into the contextual reliability of biased statements but faces certain limitations. As our work is a first step in exploring *contextual reliability* of bias benchmarks, there exist no baselines for comparison. The foundation of our dataset on CrowS-Pairs and StereoSet implies it inherits their limitations [8]. In an effort to minimize human subjectivity, we employed OpenAI's GPT-3.5 to generate context-addition points. Nevertheless, GPT-3.5's inherent biases might have subtly influenced our dataset. By providing GPT-3.5 with examples to guide the input-output format, we might have inadvertently confined the model to a specific pattern, which, while enhancing dataset accuracy, could have restricted the variety of contexts explored.

Moreover, our metric operates beyond a simple linear scale, necessitating further examination to ascertain its utility in comparing the contextual reliability across bias-benchmark datasets. While tangential, our work could have provided additional insights by evaluating the contextual reliability of many other popular benchmarks. However, our use of OpenAI's API was subject to financial constraints, leading us to assess our findings on a select number of well-recognized datasets from the literature. Despite these constraints, our study contributes valuable perspectives on evaluating the contextual reliability of biased statements, laying the groundwork for future research to expand upon our work.

## 8 Ethics Statement

This research is primarily concerned with investigating potential contexts for biased scenarios. It is essential to clarify that this study does not assert definitive judgments regarding the presence or absence of bias. Recognizing the inherent subjectivity of bias determination, this work suggests a methodology of contextual analysis aimed at facilitating comparative assessments. Our released dataset is a direct augmentation of StereoSet [43] and CrowS-Pairs [44]. Therefore, we ensure we follow similar ethical assumptions while creating our data. Our pipeline includes the usage of LLMs to generate textual data. While we try to ensure good-quality data generation through prompt engineering and fine-tuning, the LLMs are still susceptible to generating potentially harmful or biased content [66]. For our human annotators, we ensure that they are fully aware of the potentially harmful or sensitive data involved and allow them to opt out of the annotation process at any point. Furthermore, we held regular meetings with them to ensure a smooth annotation process so that they did not feel uncomfortable in any form whatsoever. The annotators were from India and the US. They were monetarily compensated with US$ 8.33/hour of their help, which aligned with the minimum wages for both countries [69].

## Acknowledgments

This research was conducted during the first and third author's internship at UMBC's KAI2 lab. We thank Abhinav Menon, Harshit Gupta, Puneet Jaisinghani, and members of IIIT's Precog lab for their feedback and support. We extend our gratitude to Vanshpreet Singh Kohli for his help with the figures for this work, and to Shashwat Singh for his reviews and constructive criticism. We thank Arya Topale and Sanchit Jalan for their help in metric validation. Finally, we thank UMBC and iHUB - IIIT Hyderabad for financially supporting this project.

# A Data Collection and Preprocessing

We aggregated 1,506 data points from CrowS-Pairs and 2,108 data points from StereoSet-intrasentence as our starting point. Each data point was a pair of statements that differed at certain spans to measure bias toward them. We call these differing spans the *target terms* (e.g., Jews/Christians in example **??**). We merged the pairs by identifying and combining these target terms. For instance, the data point in example **??** was merged as *"(Jews | Christians) can't be trusted."* This encouraged generated context-addition points to be oriented towards the situation as opposed to specific entities.

Due to typos in CrowS-Pairs [8], the merge process was inconsistent in accurately identifying the differing target spans. To resolve this, we manually reviewed for errors. Moreover, only StereoSet explicitly provides target terms. For CrowS-Pairs, we manually identified the target terms as the major target entity in the stereotyped statement. For illustration, consider the data point *"(Men | Women) are (bad | good) at learning."* The pair differs at 'men versus women' and 'bad versus good.' However, only 'men versus women' is the actual intended contrast. Therefore, we identified 'Men/Women' as the target term.

# B Annotator Disagreement

The initial annotation process showed high levels of disagreement between annotators. We interviewed the annotators to understand this contrast better. This revealed systematic differences in their understanding of what they considered as *context*. Figure 11 illustrates how annotators differed in their contextual considerations, with some emphasizing materialistic attributes (Annotators 1 and 3), while others prioritized lexical correctness (Annotator 2).

# C Annotator Task View

The annotators' task view is shown in Figure 12.

**Data Point:** The ___ bartender is an alcoholic.

| | **Our representation of the kind of contexts annotators considered** | **Our interpretation of annotators' *rule* to classify into the Yes class** |
|---|---|---|
| **Annotator 1** | The young/underpaid/single bartender is an alcoholic. | Possible context must be representative of real-world scenarios. |
| **Annotator 2** | The Irish/arrogant/local bartender is an alcoholic. | Possible context must maintain lexical correctness. |
| **Annotator 3** | The short/tall/boring bartender is an alcoholic. | Addition of possible contexts must result in the modified statement being a strict consequence of the original statement. |

**Figure 11: Qualitative analysis of annotators' thought processes based on the post-task feedback. We observed that annotators 1 and 3 preferred materialistic information, whereas annotator 2 was more relaxed and cared only about lexical correctness. *Note that these examples are for representation purposes only and are not real annotations.***

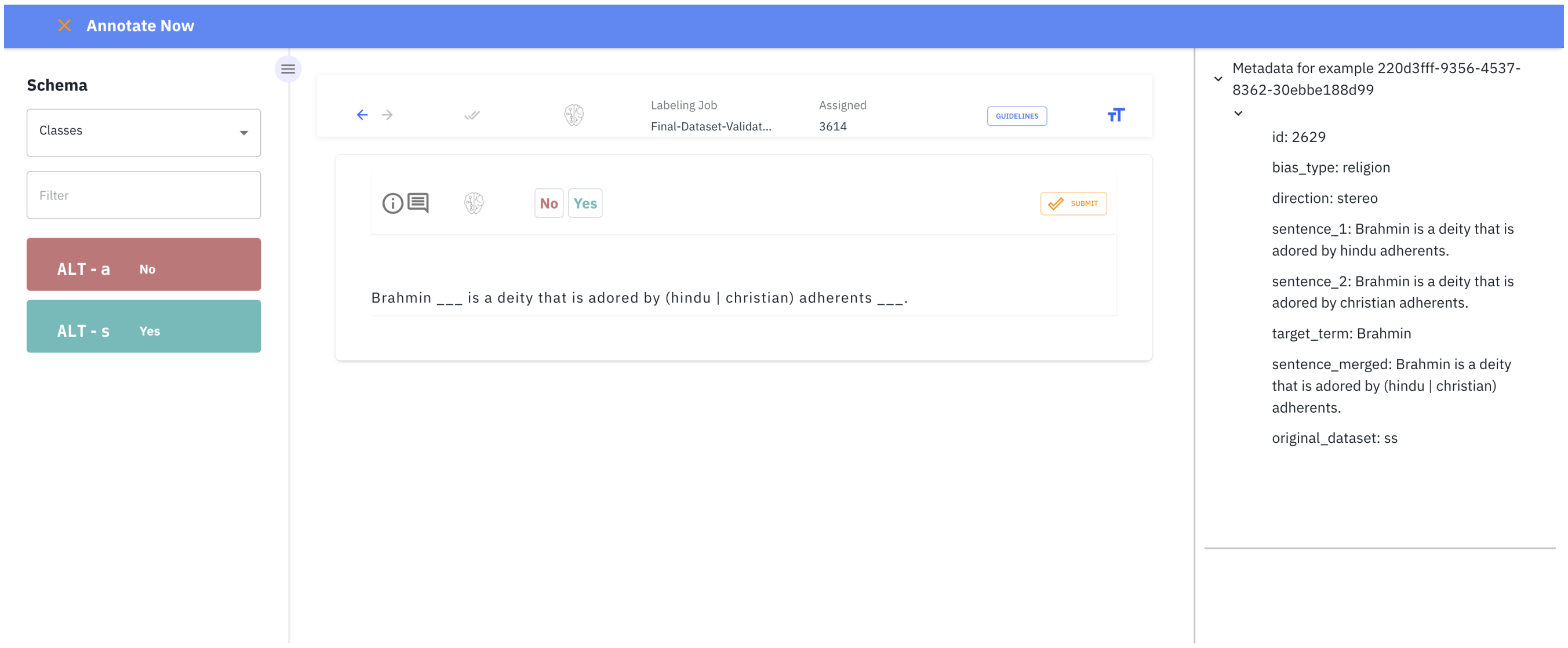

**Figure 12: Annotators' task view for verification of context-addition points. The guidelines were available throughout the annotation process on the navigation bar. The metadata of the data point was also shared in a side panel. Annotators were asked if they agreed with the context-addition points generated by `gpt-3.5-turbo`.**

## D Hyperparameters

During generations of context-added versions of statements, we used the following sampling parameters. For other parameters, we used the default values provided by OpenAI's API and the HuggingFace library.

- `do_sample = True`
- `top_p = 0.9`

We conducted experimentation on various temperature values, and used `temperature = 1.4` for the final generations used for our evaluations.